%% file: RobotDog.tex
\documentclass[conference]{IEEEtran}

\usepackage{amsmath}
\usepackage{amsfonts}
\usepackage{multirow}
\usepackage{amssymb}
\usepackage{graphicx}
\usepackage{epstopdf}
\usepackage{enumerate}
\usepackage{color}

\usepackage{booktabs, tabularx}
\usepackage{multirow}

\usepackage{epstopdf}
\usepackage{epsfig}
\usepackage{subfigure}

\usepackage{amsmath}
\usepackage{algorithm}
\usepackage{algpseudocode}

\input{macros}

\ifCLASSINFOpdf

\else

\fi

\begin{document}

\title{Humans as Path-Finders for Safe Navigation}

\author{A. Antonucci, P. Bevilacqua, S. Leonardi, L. Palopoli, D. Fontanelli

  \thanks{A. Antonucci, P. Bevilacqua, S. Leonardi, and L. Palopoli are with the Department of
	Information Engineering and Computer Science (DISI), University of
	Trento, Via Sommarive 5, Trento, Italy {\tt\small
		\{alessandro.antonucci, luigi.palopoli\}@unitn.it}.
	D. Fontanelli is with the Department of Industrial
	Engineering (DII), University of Trento, Via Sommarive 5, Trento,
	Italy {\tt\small daniele.fontanelli@unitn.it}.}
}

\IEEEoverridecommandlockouts

\maketitle
\thispagestyle{empty}
\pagestyle{empty}

\begin{abstract}
  One of the most important barriers toward a widespread use of mobile
  robots in unstructured and human populated work environments is the
  ability to plan a safe path. In this paper, we propose to delegate
  this activity to a human operator that walks in front of the robot
  marking with her/his footsteps the path to be followed. The
  implementation of this approach requires a high degree of robustness
  in locating the specific person to be followed (the {\em
    leader}). We propose a three phase approach to fulfil this goal:
  1. identification and tracking of the person in the image space,
  2. sensor fusion between camera data and laser sensors, 3. point
  interpolation with continuous curvature curves.  The approach is
  described in the paper and extensively validated with experimental
  results.
\end{abstract}
      
\input{introduction}

\input{related}

\input{tracking}

\input{navigation}

\input{experiments}

\input{conclusions}



\bibliographystyle{IEEEtran}
\bibliography{RobotDog}

\end{document}

%% file: macros.tex
\newcommand{\frm}[1]{\langle #1\rangle}



%% file: introduction.tex
\section{Introduction}
\label{sec:introduction}

When an autonomous mobile robot of remarkable size and mass navigates
the treacherous waters of unstructured and human-populated
environment, safety concerns and regulation constraints take centre stage
and 
become a barrier for the adoption of this technology.  To mitigate this problem, we advocate
a mixed approach.  When the mobile robot travels across a safe or
segregated area, it can move in full autonomy, whilst whenever it
enters a shared or dangerous area, the responsibility of the most
critical decisions (i.e., motion planning) is shifted to a human
operator.

Our reference scenario can be described as follows.  The
mobile robot starts its mission with a person standing in front.  The
robot looks at the person with its visual devices, extracts a number
of important features and elects her/him as a leader. Then starts the
second phase: the person walks to the destination, with the
robot tracking its positions and following her/him moving along the
path marked by her/his footsteps. After the leader reaches the
destination, the path is memorised and can be used
for future missions.  Observe that \emph{this is not} a standard
leader-follower application in which the robot is allowed to sway
sideways as far as it keeps a specified distance from its leader. In
our case, the human is a path-finder and the robot follows her/his
virtual footprints. The advantages are manyfold.  From
the perspective of the robot, the human acts as an external module for
the motion planning task, simplifying the complexity of the software
components and of the sensing subsystems. From the perspective of the
operator, he/she is in condition to drive a complex and
heavy robot without any skill other than being able to walk. From the
system integrator point of view, shifting the responsibility of motion
planning to a human whenever the robot moves in a difficult zone eases
the pressure of regulation and safety constraints.

The idea outlined above can be seen as an original and modern
application of the teach-by-showing approach to mobile robots moving
in a complex and dynamic scenario.  This is classified in the recent
literature~\cite{islam2019person} as a very relevant and largely open
problem and is the key methodological contribution made in this paper.
Other contributions have a more technical nature and lie in the design
of the system components.  Our processing and execution  pipeline
has three phases: 1. identification of the leader within the front camera image
frames, 2. fusion of the visual
information with the information coming from other sensors,
3. reconstruction of a smooth and feasible path from the time series
of the leader's positions, and controlled motion along the path.

The first phase is troublesome because the position of the leader is
extracted from a noise source, in which an ambiguous classification of
the different subjects in the scene is quite frequent. Our solution is
to split the first phase in three sub-phases.  The first one detects
the objects of interest within the image using a state-of-the art
convolutional neural networks (CNN) detector. The second sub-phase
recognises the leader between the objects detected in the
image. \emph{The original solution of the paper} is to first train a
neural network to classify the person and then use its last layer as a
feature set.  The feature identification is kick-started during the starting phase and
is continuously refined during the system operation.  The recognition
properly said is performed by a K-Nearest Neighbour (KNN) classifier,
which identifies the closest match with the leaders feature set
between the objects identified in the frame. The third sub-phase
consists of a tracking module, which ensures continuity in the
estimated positions of the target across different frames.

In the second phase, we fuse the information on the leader position in
the image  with the measurements of a LIDAR sensor in order the
reconstruct the correct location of the target and its headway
distance from the robot.  The resulting estimate is used as a refined
and robust measurement in a multiple model Kalman filter (KF), which
leverages two dynamic models for the prediction of human motions. The
KF is used both to increase the robustness of the estimated positions
and to track the target for some time even when he/she falls out of
the visual cone (which is an unavoidable condition when the robot has
to follow the reference footsteps in sharp turns).  The filtered
information is also fed back to the image tracking component to solve
ambiguities and avoid misclassifications.  The specific combination of
tracking filter and neural network to estimate the position of the
leader is \emph{another important technical contribution of the
  paper}.

The third phase processes the time series of the estimated position of
the leader, refining the path and guiding the navigation. This step
uses clothoid curves to interpolate the points, which produces a path
with continuous curvature and easy to follow for a robot. Finally, the
control module follows the estimated path and enforces the necessary
safety policies.

The paper is organised as follows. In Section~\ref{sec:related}, we
summarise the most important existing results that we used as
reference for this work.  In Section~\ref{sec:architecture}, we
present our general architecture and provide details on the perception
components that allow us to localise and track the leader. In
Section~\ref{sec:navigation}, we show our solution for path
reconstruction and the control strategy for following the path. The
experiments supporting the validity of the approach are described in
Section~\ref{sec:experiments}. Finally, in
Section~\ref{sec:conclusion} we give our conclusions and announce
future work directions.

%% file: related.tex
\section{Related Work}
\label{sec:related}

People following is a complex activity requiring a combination of
perception, planning, control, and interaction strategies. Following a
specific person rather than any person adds more to the complexity of
the problem and is largely classified as an open problem. The main
issue is that in a complex scenario many people can look
similar if they do not wear specific markers.  Most of the methods
developed in the last decade and surveyed by Islam et
al~\cite{islam2019person}, claim a good performance in detection and
tracking of humans, while less of one half apply online learning or
perform person re-identification, and even fewer do both.  Target
re-identification and recovery were handled first with probabilistic
models (e.g. Kalman filters), features-based techniques and more
recently with appearance-based deep networks, but these methods were
not investigated further for human-following applications. The
combination of detection, tracking, and recognition was proposed by
Jianng et al.~\cite{jiang2018long} using the Speeded Up Robust
Features (SURF). However, the key point matching did not show a
sufficient level of robustness in our experimental tests with human
figures.  The complexity of the problem requires the combination of
sophisticated learning approaches, model based filtering and path
interpolation, as shown in this paper.
    
\noindent {\bf Object Detection.} Object detection is in our framework
the first element of the processing pipeline. For this component, we
sought a good compromise between classification accuracy and
achievable frame rate. The available methods range from object
detection and segmentation methods~\cite{liu2016ssd,
  redmon2016you,girshick2014rich}, to specific solutions for human
pose detection~\cite{cao2019openpose}. YOLO~\cite{redmon2016you} is a
very effective solution based on a single CNN; its main known
disadvantage materialises when two classes have similar probabilities
or the shape of the element is not perfect and the algorithm could
produce different bounding boxes for the same object. Alternative
solutions such as SSD~\cite{liu2016ssd} apply correction techniques to
overcome the limitation~\cite{neubeck2006efficient}. After a thorough
performance assessment, we evaluated that SSD was the best compromise
for our application.

\noindent {\bf People recognition.}  People recognition in computer
vision is difficult in its own right.  An additional level of
complexity is introduced by the fact that the camera used for image
acquisition is mobile.  Traditional offline algorithms like Support
Vector Machines (SVM)~\cite{hearst1998support} are known to react
quickly to classification queries, but are not a good fit for our
scenario, because we lack a prior knowledge on who is going to be the
leader and we need to be robust against possible changes in her/his
appearance.  Methods based on feature point
matching~\cite{pun2015image} are known to be robust and are widely
used to find small patterns in complex images, but in our tests the
PRID450 (Person Re-IDentification) dataset~\cite{roth14a} showed
a high number of errors for low-res images and for deformable shapes
such as humans clothes.  Our final solution was based on the use of a
K-Nearest Neighbours (KNN) classifier, which is an efficient
training-free classification method. However, it requires the
knowledge of representative points for the classification. For this
information we used the last layer of a CNN, which gets trained with
the different views of the leader.
The idea of using a CNN classifier to extract the feature set was
presented by Ristani et al.  in~\cite{ristani2018features}, who
proposed this idea to match detections from multiple cameras.  The
classifiers evaluated for this work are the Deep Neural Networks
(DNNs) based GoogLeNet~\cite{szegedy2015going} and
ResNet~\cite{he2016identity}.
GoogLeNet
showed the best performance in our experiments.

\noindent {\bf Person Tracking in the video frames. } For person
tracking, we could select from a large variety of approaches for the
tracking of general objects (the fact that our object of interest is a
person does not make a big difference in this case).
Specifically, we considered: the Multiple Instance
Learning (MIL) tracker~\cite{babenko2010robust}, the Kernelised
Correlation Filters (KCF) tracker~\cite{henriques2012exploiting}, the
Median Flow tracker~\cite{kalal2010forward}, the Channel and Spatial
Reliability Tracker (CSRT)~\cite{lukezic2017discriminative}, the
Minimum Output Sum of Squared Error (MOSSE)
tracker~\cite{bolme2010visual}, the Generic Object Tracking Using
Regression Networks (GOTURN) tracker~\cite{held2016learning}, and the
Tracking-Learning-Detection (TLD)~\cite{kalal2011tracking}. 
After a thorough performance analysis, partly reported in Section~\ref{sec:experiments}, we selected CSRT as the most suitable solution for our purposes.

\noindent{\bf Sensor fusion.}  Our application requires 3D
reconstruction of the human pose. The combination between Stereo and
RGB-D sensor with skeleton-based approaches proves very useful to this
purpose and it is significantly simplified by the availability of
public domain software
components~\cite{antonucci2019performance}. However, the simple use of
visual information has known limitations such as the sensitivity to
lighting conditions, and the high computation times.  Laser-based
sensors, on the other hand, are relatively reliable on a long range
and are less computation hungry than vision based approaches.  However,
recognising a specific person from a slice of a 2D point cloud is
hopeless. For this reason, moving along a direction frequently taken
in robotics~\cite{zhen2019joint,wolcott2014visual,hwang2016fast}, we
apply a combination of cameras and LIDARS.  The use of separate
systems for depth estimation and classification improves the robustness
of the tracking system when one of the sensors fails; e.g., if the
leader falls outside the camera field of view, we can still use the
LIDAR sensor for some time assuming a certain degree of reliability of
the specific-person following.

%% file: tracking.tex
\section{Tracking the human path-finder}
\label{sec:architecture}

\begin{figure}[t]
  \centering \includegraphics[width=\columnwidth]{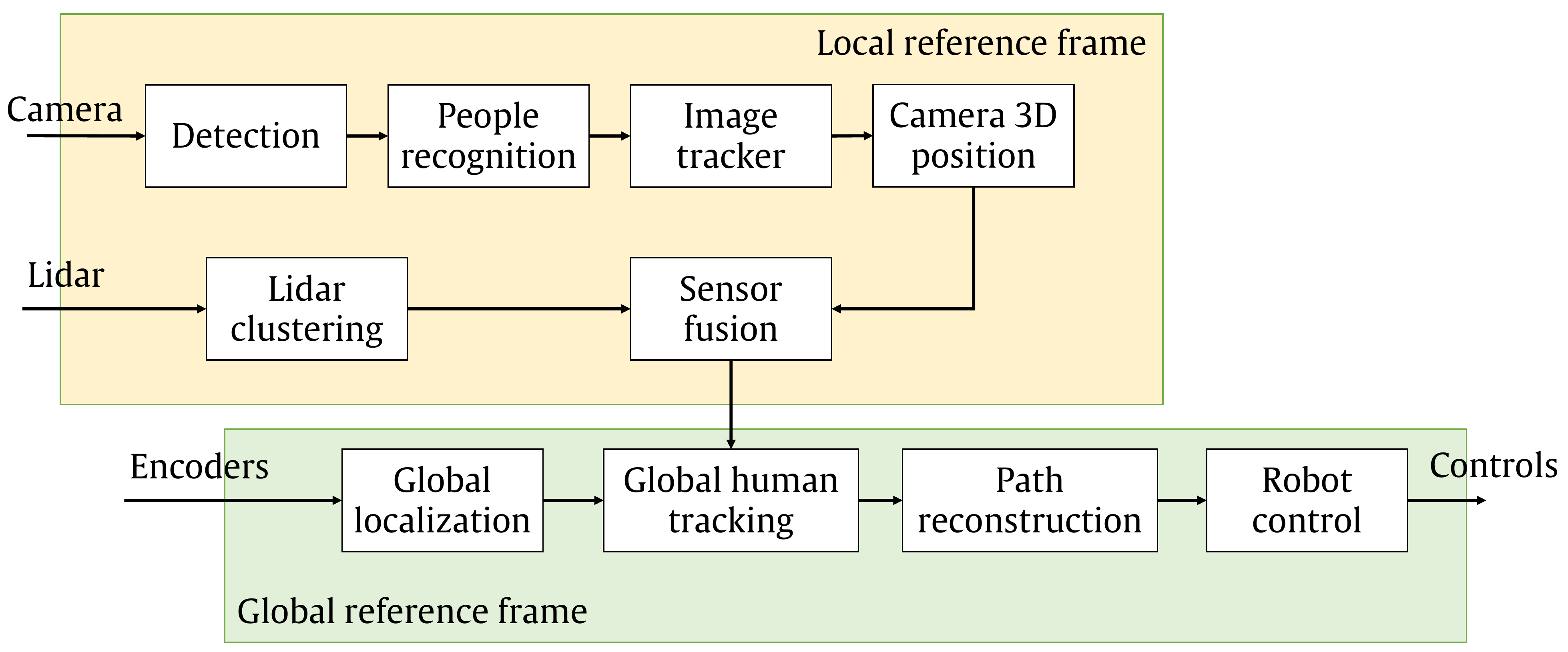}
  \caption{Overall scheme of the algorithm.}
  \label{fig:scheme}
\end{figure}

The proposed algorithmic framework is sketched in
Figure~\ref{fig:scheme}.
Before going into the details of the different building blocks, we
succinctly describe the available sensing system and the model of the
platform. The reference model for the robot is in this paper the unicycle,
which can be described in discrete time by the following kinematic model:
\begin{equation}
  \label{eq:DiscrModelUni}
  s(t_{k+1}) = 
  \begin{bmatrix}
    x_r(t_k) + \cos(\varphi_r(t_k)) (t_{k+1} - t_{k}) v_r(t_k) \\
    y_r(t_k) + \sin(\varphi_r(t_k)) (t_{k+1} - t_{k}) v_r(t_k) \\
    \varphi_r(t_k) + (t_{k+1} - t_{k}) \omega_r(t_k)
  \end{bmatrix}
\end{equation}
where $s(t_{k}) = [x_r(t_k), y_r(t_k), \varphi_r(t_k)]^T$ is the state
of the robot, the Cartesian coordinates $(x_r(t_k), y_r(t_k))$ refer
to the mid-point of the rear wheels axle in the $X_w \times Y_w$ plane
expressed in the $\frm{W}=\{X_w, Y_w, Z_w\}$ world reference frame,
$\varphi_r(t_k)$ the longitudinal direction of the vehicle with
respect to the $X_w$ axis, $v_r(T_k)$ and $\omega_r(T_k)$ the
longitudinal and angular velocities, respectively, and $t_k$ the
reference time instant, which is usually chosen an integer multiple of a fixed sampling time. Importantly, the proposed framework would be applicable
to different robot dynamics; however, as explained next, the unicycle
structure is particularly convenient for the class of applications we address.

Without loss of generality, we assume here that the choice of the
sampling time $\delta_t = t_{k+1} - t_{k}$ is imposed by the sensor
with the lowest sampling frequency.  The assumed sensing configuration
is based on the presence of rotation encoders on each of the rear wheels
or any other sensing system able to provide ego-motion informations
(e.g., IMUs, visual odometry). For the perception of the surroundings,
the sensing system comprises a LIDAR and an RGB-D camera. The LIDAR
data are used to both track human beings around the vehicle and
to localise the vehicle inside the environment.  The RGB-D camera is
primarily used for the human detection and tracking. The laser scanner (an
RPLidar A3\footnote{https://www.slamtec.com/en/Lidar/A3}) employed has
a view of $360^\circ$, a maximum measuring distance up to 40 meters, and is typically operated at 20 revolutions per second. The RGB-D camera adopted is an
Intel\textregistered{} RealSense\texttrademark{}
D435\footnote{https://www.intelrealsense.com/depth-camera-d435/},
working in an ideal range spanning from $0.5$ to $3$~m, whose images
are adopted in the vision-based detection and recognition system described in 
Section~\ref{subsec:camera-tracking}.

\begin{figure}[t]
	\centering
	\begin{tabular}{cc}
	\includegraphics[width=0.3\columnwidth]{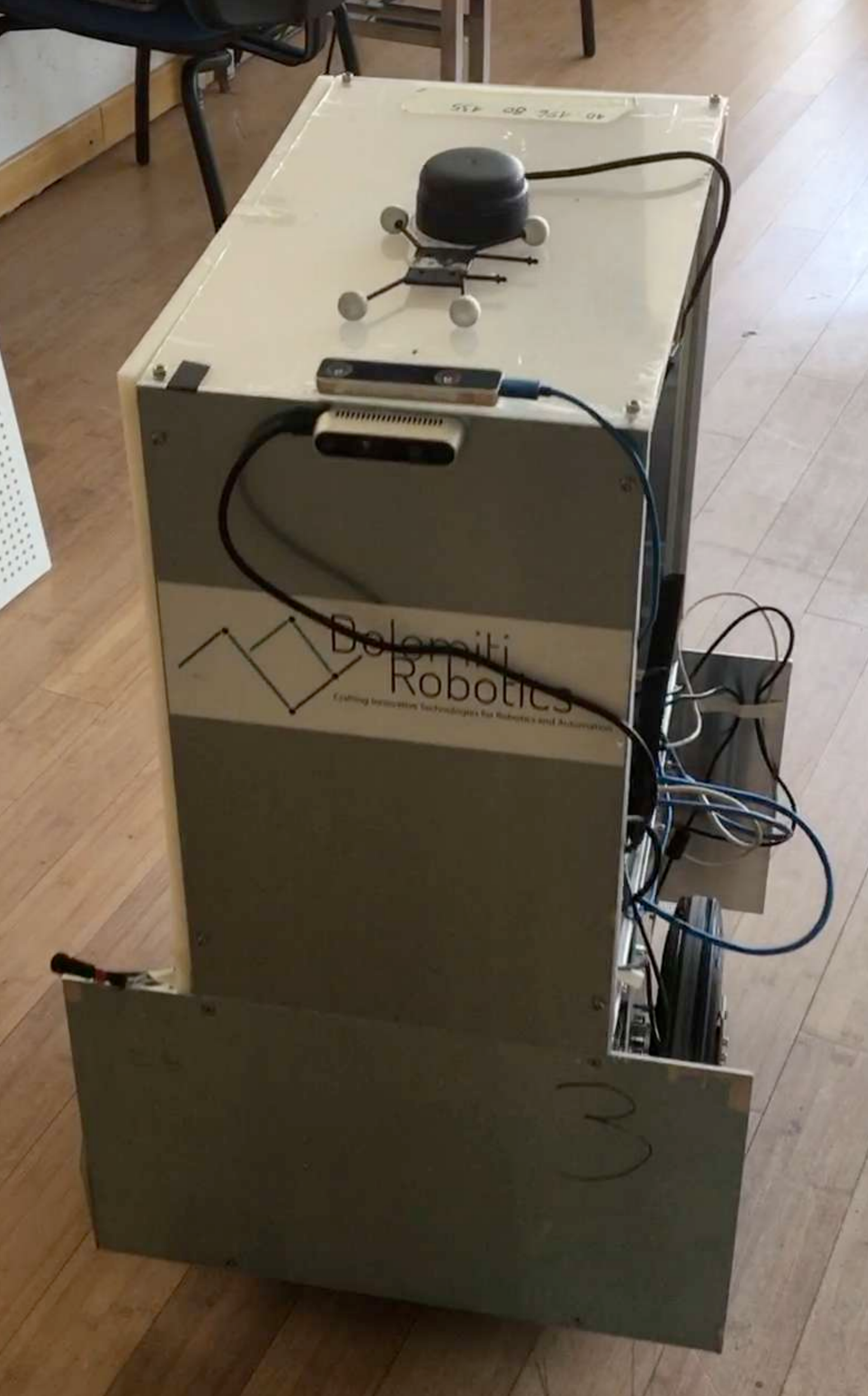} &
	\includegraphics[width=0.48\columnwidth]{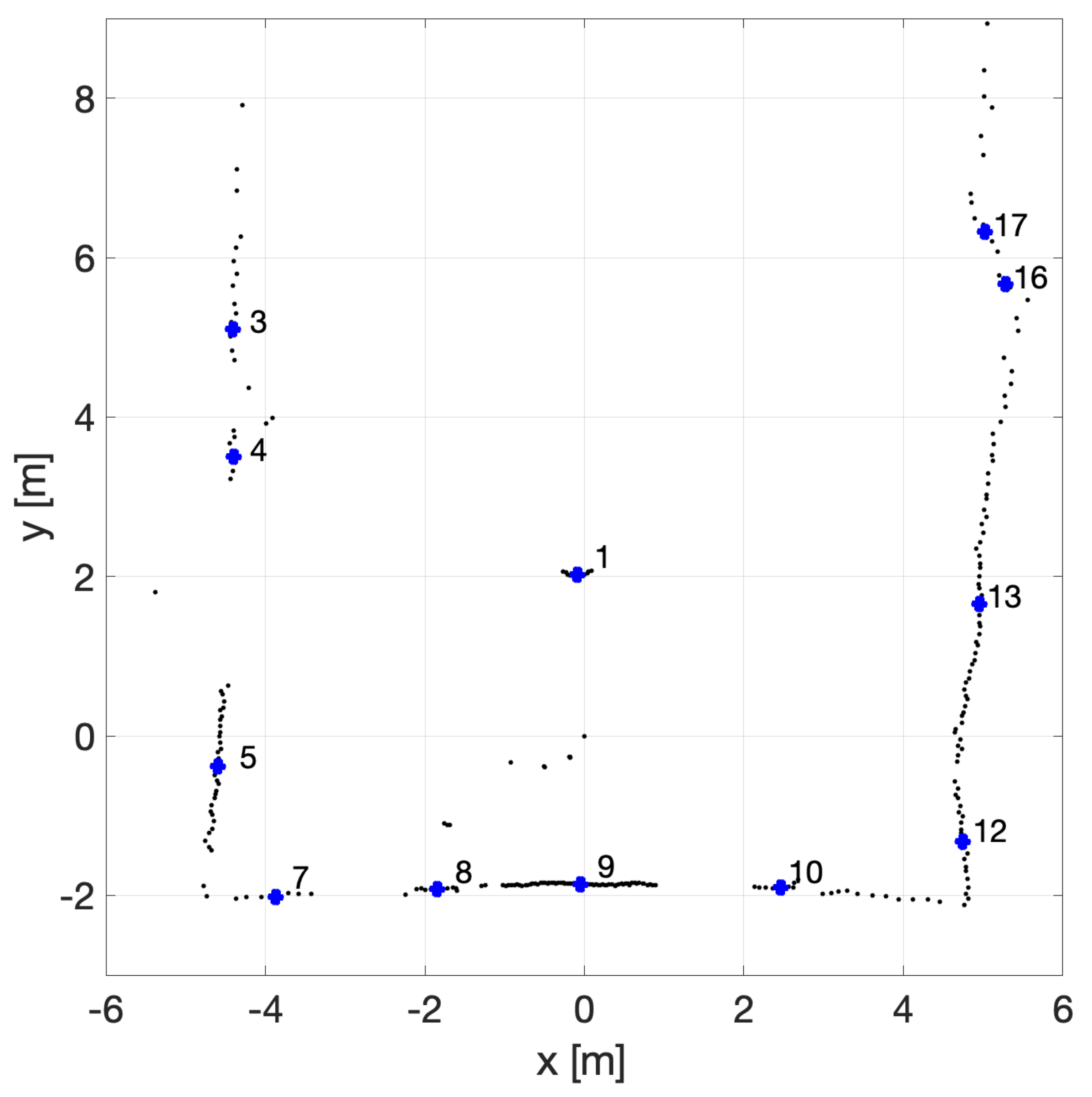} \\
	(a) & (b) \\
	\end{tabular}
	\caption{(a) Robot sensing system setup, consisting of LIDAR
          sensor, RealSense D435, and RealSense T265 (for the visual
          odometry). (b) Laser scanned cloud points (thin black dots),
          with the object centroids (thick blue points) expressed in
          the LIDAR reference system $\frm{L}$.}
	\label{fig:lidar-robot}
\end{figure}

The LIDAR and the camera are rigidly mounted on the top of the robot
chassis (see Figure~\ref{fig:lidar-robot}-a) and return the measurements in the
reference frames $\frm{L}$ and $\frm{C}$, respectively, which are bot
rigidly linked to the robot (i.e., they operate with a local coordinates
reference system).
The transformation matrix $^{L}T_{C}$ between the two frames is estimated
during the calibration phase in order to simplify sensor fusion.

\subsection{Vision-based detection and recognition}
\label{subsec:camera-tracking}

The detection and recognition algorithm is almost entirely based on
the images grabbed by the camera and comprises the detection, the
recognition and image tracking phases. However, the information
retrieved from the camera is interleaved with the one retrieved from the LIDAR (see Sec.~\ref{subsec:lidar-fusion}).

{\bf Detection}: The vision-based detection module is used to
periodically localize $D$ people inside an image frame. Since we have
chosen the latest version of YOLO available,
YOLOv3~\cite{redmon2018yolov3}, and a lighter implementation of SDD,
namely MobileNet~\cite{howard2017mobilenets} (designed to execute on
low power devices), the detection module models the objects in view as
the smallest bounding box that contains the detected element. In the
starting phase, the person associated with the largest bounding box is
recognised as leader.

\begin{figure}[t]
  \centering
  \begin{tabular}{cc}
    \includegraphics[width=0.48\columnwidth]{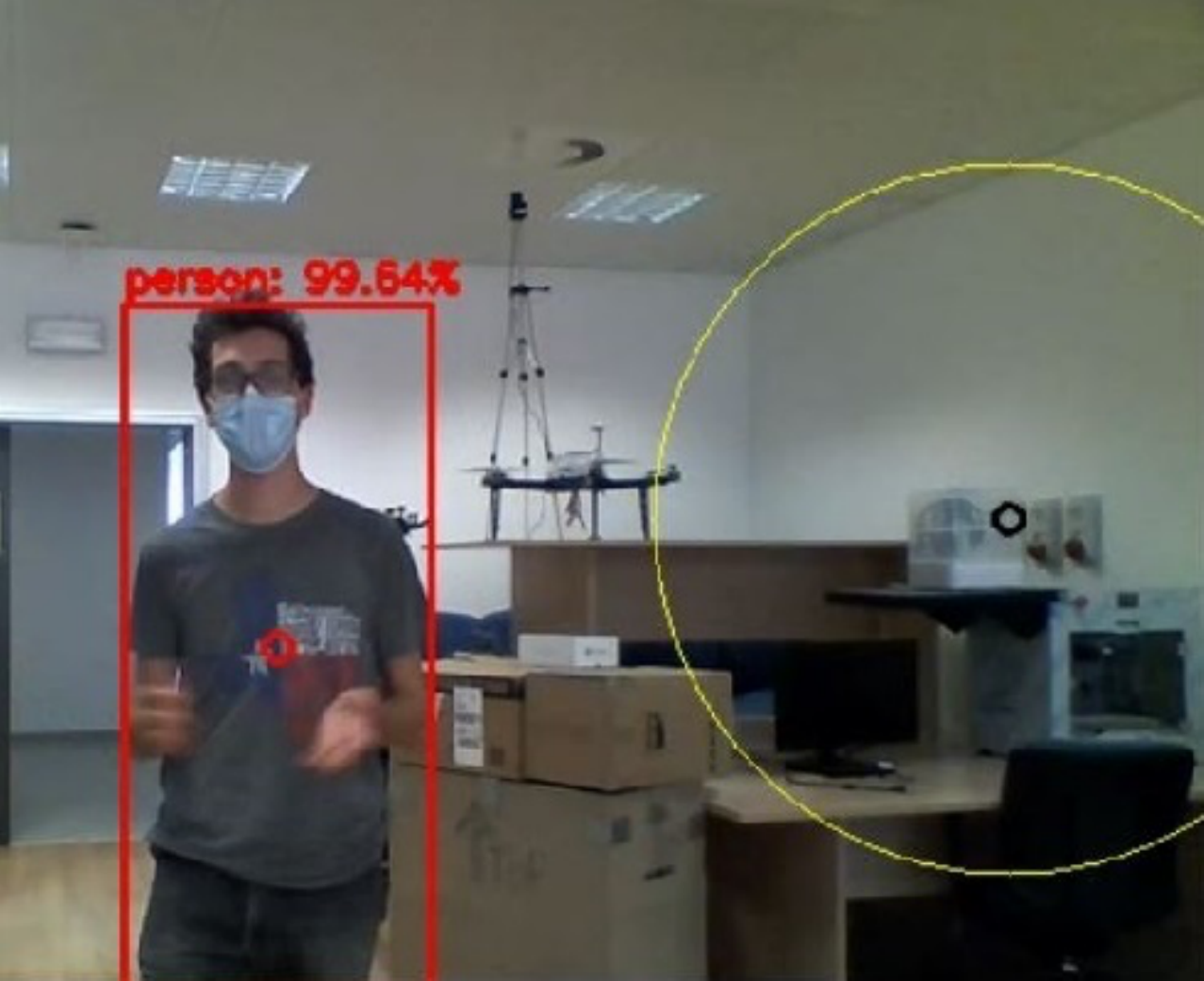} &
                                                                            \includegraphics[width=0.38\columnwidth]{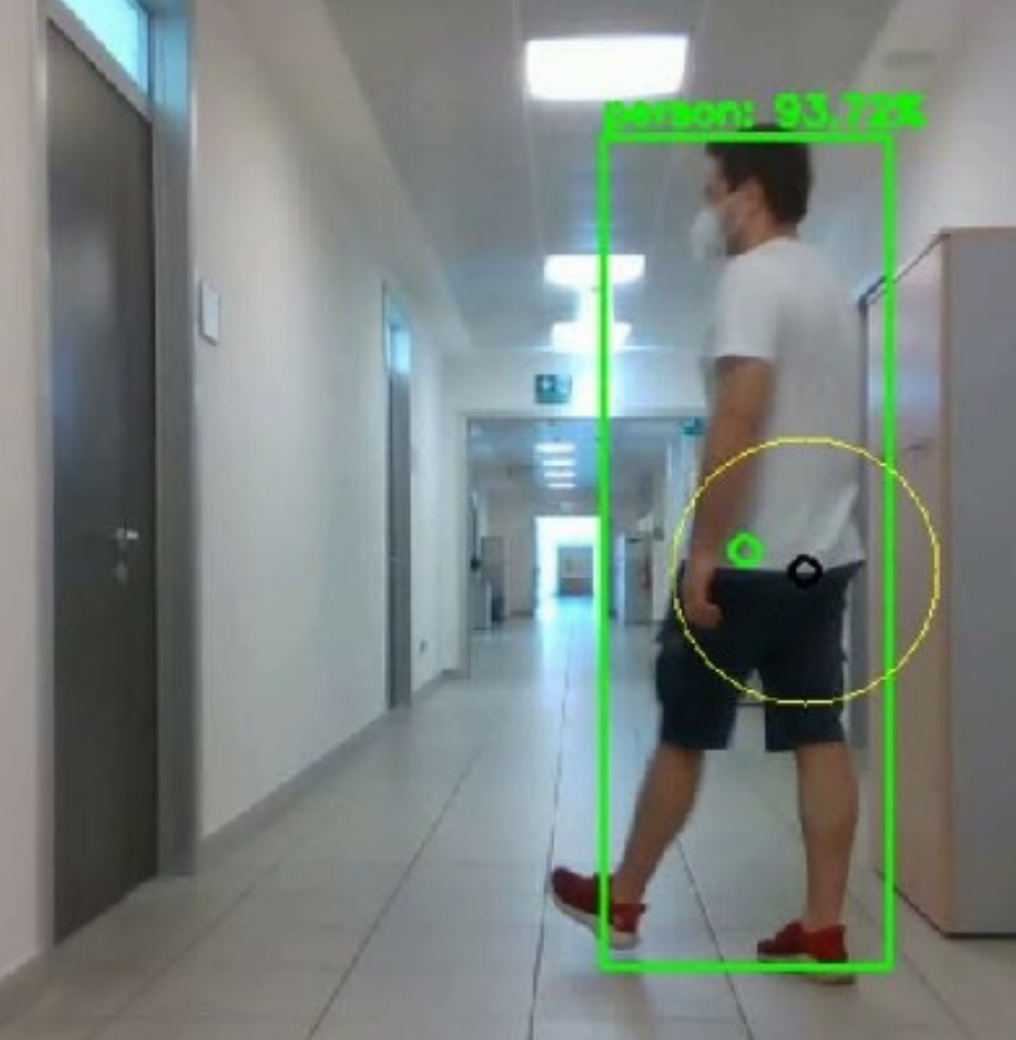} \\
    (a) & (b)\\
  \end{tabular}
  \caption{Two examples of the application of the drift
    tolerance~\eqref{eq:drift}. In (a), the person is correctly
    classified as a mismatch because the bounding box is too far apart
    from the region defined by~\eqref{eq:drift}, while in (b) the
    leader is newly detected as expected.}
  \label{fig:drift}
\end{figure}

Since the estimation drift affects all the tracking algorithms (and
becomes particularly dangerous on long video sequences), we have
assumed a proximity constraint along sequential frames. More
precisely, during subsequent detection phases, if the detected target
is too far away from the last known position of the leader, the detected shape is ignored and the information on the leader position is recovered from the LIDAR. The
tolerance is expressed as a circle centred in the centroid of the last
valid bounding box and whose radius is defined as
\begin{equation}
d = t s \left(\ \dfrac{w}{100} \right)^2,
\label{eq:drift}
\end{equation}
where $t$ is the time elapsed from the last correct detection of the
leader, $s$ is a tuning parameter empirically set as $0.05$ and $w$ is
the width of the last bounding box from the image tracker, used to
simulate the distance of the leader from the camera and based on the
fused information of Section~\ref{subsec:lidar-fusion}.
Figure~\ref{fig:drift} shows an example where a person is immediately
classified as a mismatch or a correct match.

\textbf{People recognition}: This module is used to understand if the
found person corresponds to the leader or not, which enables a coherent
connection between the detector and the tracker. Our method is based
on a KNN algorithm.The KNN classifier uses images converted into points with DNN
image classifiers: the ResNet50~\cite{he2016identity}, which produces
a representation point in $2048$ dimensions, and
GoogLeNet~\cite{szegedy2015going} that produces a representation point
in $1024$ dimensions.  If the leader is contained in the list of the $D$
people found, the information is passed to the image tracker,
otherwise the procedure loops the detection again.

As previously mentioned, the proposed recognition algorithm has a
major distinctive feature: during the execution of the task, the
generated representative points of the leader are fed back into the
KNN. While this makes the leader recognition very robust in crowded
spaces, a leader classified as a mismatch in the detection phase,
cannot be re-evaluated as a valid leader and, hence, cannot be
converted in a correct prediction. Albeit this unfortunate event may
always happens in dynamic environments, the presence of the fusion
algorithm in Section~\ref{subsec:lidar-fusion}  mitigates its effects. Indeed, the results of the detection phase are fused with
the LIDAR data before the information is fed into the neural
network. However, if the estimation error of the human tracking filter
in Section~\ref{subsec:HumanTrackingFilter} exceeds the desired leader
tracking uncertainty due to repetitive sensor or detection failures,
the system reaches a faulty condition, the robot stops and the process
should be reinitialised from scratch.

\textbf{Image tracker}: This module is periodically executed to track
the leader location in a fixed number of $m$ frames to avoid the
problems generated due to long-term sequences. After that, the
detection is performed again in order to strengthen the tracking
performance. To this end, we implemented the methods that best fitted
our requirements, i.e. KCF, CSRT and MOSSE. We emphasise that if a
single detection fails or the leader is not found, the tracker cannot
be started.

\begin{figure}[t]
  \centering
  \begin{tabular}{cc}
    \includegraphics[width=0.45\columnwidth]{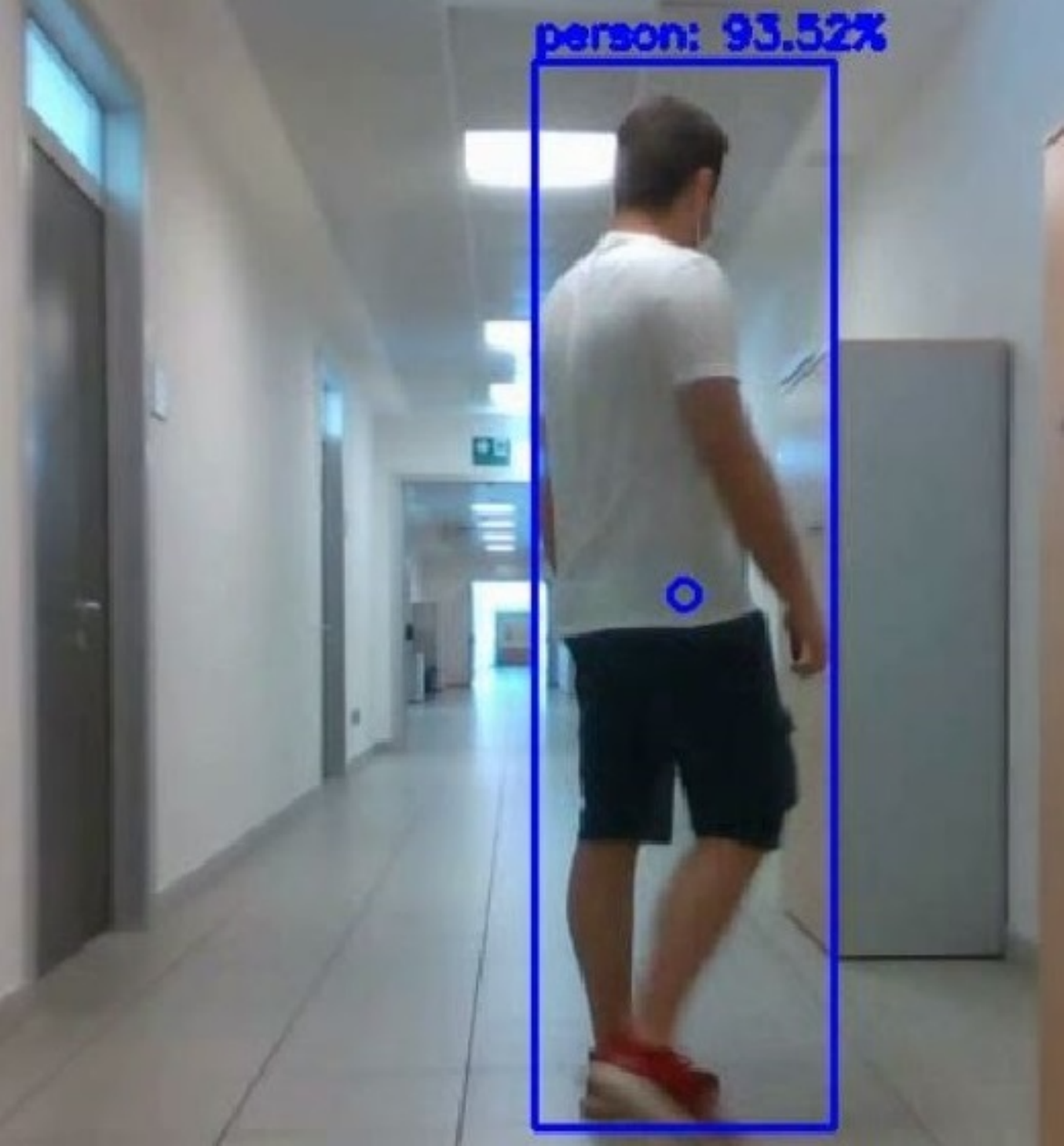} &
                                                                           \includegraphics[width=0.45\columnwidth]{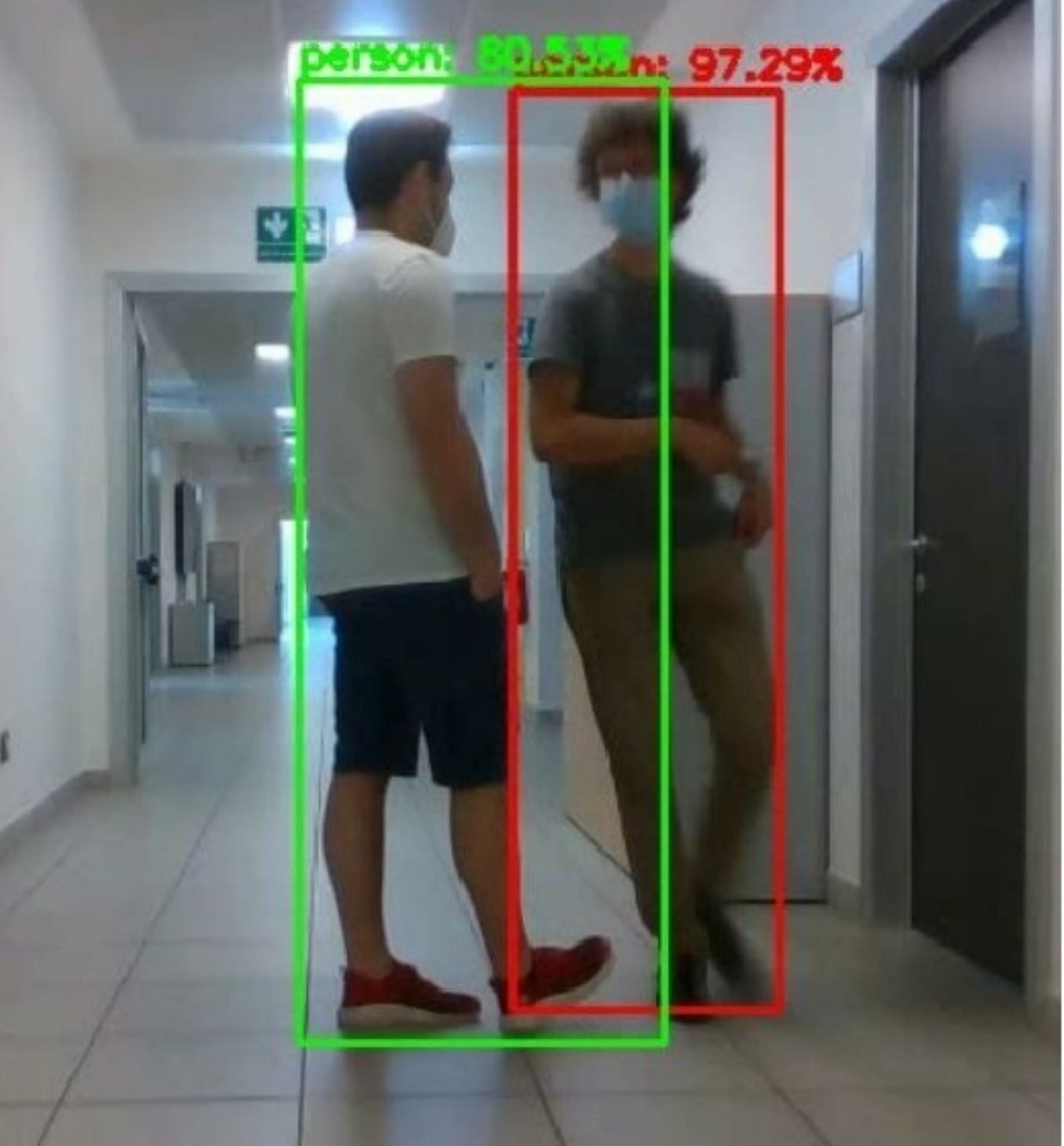} \\ (a) & (b) \\
  \end{tabular}
  \caption{(a) Initialisation phase: the detected leader is
    depicted with a blue rectangle. (b) Following phase: the leader is
    correctly recognised (green rectangle), while another person is a
    negative sample (red rectangle).}
  \label{fig:initialization-following}
\end{figure}

{\bf Overall detection and recognition algorithm}: Since the image
tracker pipeline performs an online learning classification of the
human, we start the pipeline with an \textit{initialisation phase},
where only the detection module is working. This phase is designed in order to quickly
train the KNN with a prior knowledge of the leader that will be
enforced with the previously described feedback mechanism in the
\textit{robot following phase}. In the initialisation phase, which
lasts for $\Delta_t$ seconds, the robot collects a series of bounding
boxes used to create the set of positive representative points into
the $N$-dimensional space of the KNN. Simultaneously, a negative
sample is randomly picked up from a database and it is also given to
the KNN to balance the number of positive and negative samples. The
negative samples come from our customised version of the Market1501
dataset~\cite{zheng2015scalable}.  An example of the initialisation
phase is shown in Figure~\ref{fig:initialization-following}-a.
In the following phase, a first step of detection is carried out and
all the found targets are checked with the drift-compensation
threshold of~\eqref{eq:drift}. Then, the KNN classifier is used to
distinguish between positive detections and negative ones (which are
added to the negative dataset). In the successive $m$ frames the image
tracker is executed, and the result is passed to the fusion
module. This workflow delivers a very high image
processing performance and an improved robot localisation accuracy,
as shown by the experimental data in Sec.~\ref{sec:experiments}.

\subsection{Local LIDAR-camera sensor fusion}
\label{subsec:lidar-fusion}

The information coming from the vision-based algorithm are combined
with the LIDAR in a local reference frame in order to make the
procedure more robust, as aforementioned. An additional benefit of this
phase is that the leader can be followed for some time also when he/she evades the vision cone of the RGB/D camera relying on the LIDAR information.
The sensed data are fused in
local coordinates, as explained next. 

\textbf{LIDAR clustering}: Each scan delivered by the laser scanner
provides a sequence of $n$ measurement points in the form of
$\mathcal{P} = \{p_1,\dots,p_n\}$, represented in polar coordinates as
$p_i = (r_i, \alpha_i)$, i.e. a range and an angle expressed in the
planar LIDAR reference frame $\frm{L}$, or in Cartesian coordinates,
i.e.  $p_i = (x_i, y_i) = (r_i \cos \alpha_i, r_i \sin \alpha_i)$
again in $\frm{L}$ (see Figure~\ref{fig:lidar-robot}-b for an example
of an actual scan).  At time $t_k$, the measured points are filtered
and grouped into $m_k$ clusters based on the mutual Euclidean
distances and on the richness, i.e. on a minimum number of sensed
points for each cluster, each identified by the object centroid
$o_j(t_k) = [x_j(t_k), y_j(t_k), 0]^T$, $j = 1,\dots, m_k$, and expressed
in local coordinates, i.e. in $\frm{L_k}$.

\textbf{Camera 3D position}: The tracking module described in
Section~\ref{subsec:camera-tracking} returns a bounding
box~$\left[x,y,w,h\right]$ in the image frame, containing the
$\left(x,y\right)$ pixel coordinates of the top-left corner of the
box, its width $w$ and height $h$, which is then converted in the
$\frm{C} = \left\{X_c,Y_c,Z_c\right\}$ pin-hole camera reference
system.  Notice that the depth information along the $Z_c$ axis is
retrieved via the RealSense\texttrademark API. As a consequence, the
centroid of the $i$-th bounding box
$c_i(t_k) = [x_c(t_k), y_c(t_k), z_c(t_k)]^T$ can be expressed in the
camera reference system $\frm{C_k}$ at time $t_k$.

\textbf{Sensor fusion}: Given the set of objects
$\mathcal{O}_k = \{o_1(t_k),\dots,o_{m_k}(t_k)\}$ and the centroid(s)
of the bounding box $c_j(t_k)$, taken at the same time instant $t_k$,
we adopt a spatio-temporal correspondence algorithm with the two sets
of measurements to decide if the tracked object is the same or a new
one has entered into the scene.  This simple algorithm, which adds a
physical inertia to tracked objects, filters out spurious detections
and it is implemented as a finite state machine (which is not detailed
here for the sake of brevity). The rationale is the following: let us
assume that a correct match between the $j$-th clustered object
$o_j(t_{k-1})$ and the $i$-th bounding box centroid $c_i(t_{k-1})$ is
available in the local LIDAR frame $\frm{L_{k-1}}$ (this is obtained
in $\frm{L_{0}}$ with an initialisation phase, where the leader stands
in front of the robot for about $5$ seconds for the initial
bootstrap). The robot then moves for $\delta_t$ seconds according to
the model~\eqref{eq:DiscrModelUni} and updates  its position
$s(t_k)$ in $\frm{L_{k-1}}$ either by using the encoders (if $\delta_t$ is
sufficiently small) or the global localisation module. After the
motion, the information $o_j(t_{k-1})$ and $c_i(t_{k-1})$ are
projected in new local frame $\frm{L_{k}}$. Such information form a prior
for the next leader detection, and are fused with the new sets
of measurements $\mathcal{O}_k$ and $c_i(t_{k})$, $\forall i$. Notice
that this procedure reduces at the same time the computation times and the probability of mismatch, while making the algorithm robust to lost measurements
 (either the bounding box or the LIDAR cluster are sufficient for
recognition).

\subsection{Global human tracking}
\label{subsec:HumanTrackingFilter}

Since the human is used as a path-finder for future executions of the
path, its position should be estimated in the global reference frame
$\frm{W}$. To this end, we first need to estimate the robot position
$s(t_k)$ in $\frm{W}$. This is accomplished fusing together the
encoder readings and the LIDAR points $p_i(t_k)$ with an a-priori map
of the environment. The $s(t_k)$ robot position and the leader local
measurements in $\frm{L_{k}}$ are used to obtain the Cartesian
coordinates $(x(t_k), y(t_k))$ of the leader in $\frm{W}$.

To track the human being in $\frm{W}$, an estimation algorithm is
needed, whose main role is to further improve the accuracy of the
reconstructed path and to further increase the robustness to
occasional sensor failures.  The underlying
physical assumption of the leader tracking algorithm is that the
leader moves following an unimodal probability density function, i.e.,
the human cannot move simultaneously in more than one position. In
order to limit the computational cost and to comply with the established
literature techniques, we assume that this pdf is Gaussian. Thus, two
Kalman Filters (KFs) are adopted as tracking filters, whose difference
is the adopted motion models. The first one is the constant velocity
model, a simplified version of the quite known Social Force Model
(SFM)~\cite{helbing1995social}.  In this case, a human being is
modelled as a point moving with constant velocity.  Therefore, being
$h(t_k) = [x(t_k), y(t_k), v_x(t_k), v_y(t_k)]^T$ the state at time
$t_k$ comprising the position and the velocity of the human on the
plane of motion, we have for the corresponding KF
\begin{equation}
  \begin{aligned}
    h(t_{k+1}) & =
    \begin{bmatrix}
      1 & 0 & \delta_t & 0 \\
      0 & 1 & 0 & \delta_t \\
      0 & 0 & 1 & 0 \\
      0 & 0 & 0 & 1
    \end{bmatrix} \begin{bmatrix} x(t_k) \\ y(t_k) \\ v_x(t_k) \\
      v_y(t_k)
    \end{bmatrix}
    + \begin{bmatrix}
      \frac{\delta_t^2}{2} & 0 \\
      0 & \frac{\delta_t^2}{2} \\
      \delta_t & 0 \\
      0 & \delta_t
    \end{bmatrix} \begin{bmatrix} \nu_x(t_k) \\ \nu_y(t_k)
    \end{bmatrix} = \\
    & = A h(t_k) + B \nu(t_k) ,
  \end{aligned}
  \label{eq:Kf}
\end{equation}
where $\nu(k)$ is the acceleration noise affecting the velocity
variations, supposed to be $\nu(t_k)\sim\mathcal{N}(0, Q)$, with $Q$
being its covariance matrix. The random walk hypothesis is due to the
fact that the robot has no knowledge about the actual motion
intentions of the human.

The second model instead assumes that humans actually move with a
smooth dynamic, as observed
in~\cite{arechavaleta2008nonholonomic}. Hence, the motion model can be approximated by a unicycle dynamic~\cite{farina2017walking}.  Hence, we
explicitly express the angular and linear velocities as states, and by
denoting with
$\bar h(t_k) = [x(t_k), y(t_k), \theta(t_k), v(t_k), \omega(t_k)]^T$
the state at time $t_k$ (where $\omega(t_k)$ is the angular velocity),
we have the following Extended Kalman Filter (EKF) prediction model
\begin{equation}
  \begin{aligned}
    \bar h(t_{k+1}) & =
    \begin{bmatrix}
      x(t_k) + \delta_t v(t_k) \cos(\theta(t_k)) \\
      y(t_k) + \delta_t v(t_k) \sin(\theta(t_k)) \\
      \theta(t_k) + \delta_t \omega(t_k) \\
      v(t_k) \\
      \omega(t_k)
    \end{bmatrix} + \begin{bmatrix}
      0 & 0 \\
      0 & 0 \\
      0 & 0 \\
      \delta_t & 0 \\
      0 & \delta_t
    \end{bmatrix} \begin{bmatrix} \eta_a(t_k) \\ \eta_\omega(t_k)
    \end{bmatrix} = \\
    & = f(\bar h(t_k)) + B \eta(t_k) ,
  \end{aligned}
  \label{eq:Ekf2}
\end{equation}
where $\eta(t_k)$ is the acceleration noise affecting the linear and
the angular velocities that is assumed to be
$\eta(t_k)\sim\mathcal{N}(0, E)$, with $E$ being its covariance
matrix.

The motion models are then selected using the Multiple Model Approach
(MMA) presented in~\cite{Shalom01}, which relies on the measurements
maximum likelihood approach in a Bayesian setting to estimate the
probability in being in one of the two models. However, since humans
behave differently in different situations and depending on the
context, potential mode changes are considered using the first-order
generalised pseudo-Bayesian estimator~\cite{Shalom01}.  This approach
fuses together the estimates of each model in a single estimate before
the models are adopted, and hence uses the leader measurements to both
refine the estimates and the probabilities of each of the two models.

%% file: navigation.tex
\section{Navigation}
\label{sec:navigation}

The aims of the navigation module are twofold: reconstructing the path
followed by the leader in a form that can be followed by the robot,
controlling the motion in order for the robot to follow the path with
a good accuracy (small deviations are inevitable but they should be
kept in check).
\begin{figure}[t]
  \centering \includegraphics[width=0.7\columnwidth]{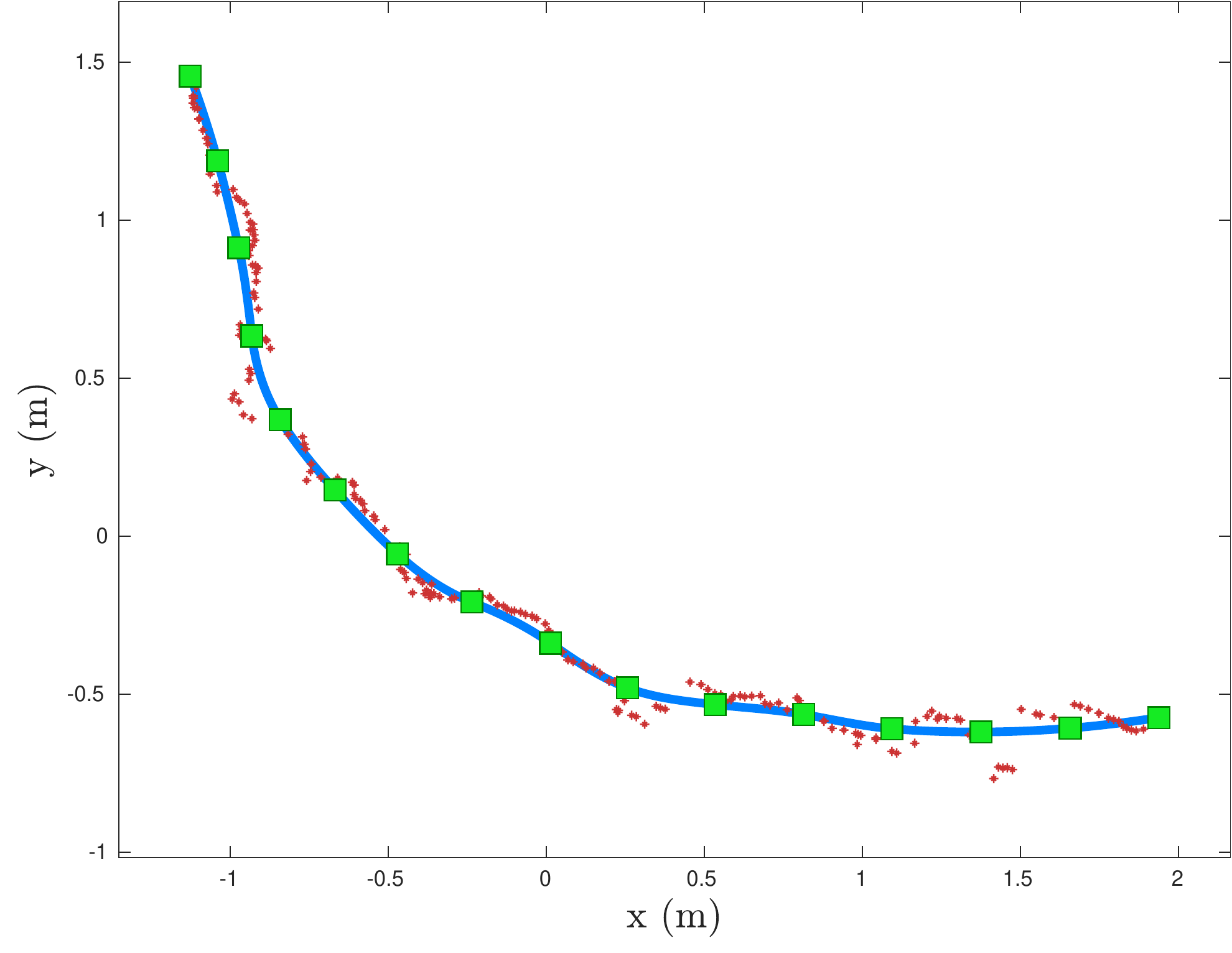}
  \caption{Example of path fitting and reconstruction. The red stars
    represent the input data.  The green squares are the fitted
    waypoints, sampled at a uniform distance along the path.  The blue
    solid line is the reconstructed, smoothed path, to be followed by
    the robot.}
  \label{fig:pathFitting}
\end{figure}

\subsection{Path reconstruction}
As shown in the scheme in Figure~\ref{fig:scheme}, the path
reconstruction module continuously receives new information on the
current position of the leader from the perception module.  This way,
it creates a dataset composed of a time series of 2D leader positions
which are updated in real--time. The module executes a local path
fitting of the estimated leader trajectory. An example execution of the process is shown in  Figure~\ref{fig:pathFitting}.

The path is reconstructed using the following steps.
\begin{enumerate}
\item Once a new leader position is received, it is compared with the
previous one, and, if the distance is greater than a small, threshold
value, it is recorded into the dataset. This action is necessary to
handle the scenario where the leader stops for a long time, in order
to avoid an unnecessary growth of the dataset.
\item When the new position
qualifies for its inclusion into the dataset, the $x$ and $y$
components of the datapoints are fitted using a classical smoothing
algorithm, i.e. the LOESS (Locally Estimated Scatterplot
Smoothing)~\cite{cleveland1979robust}.
\item The fitted datapoints are then
connected by a polyline, and a number of waypoints are sampled at a
uniform curvilinear distance (corresponding to the green squares of
Figure~\ref{fig:pathFitting}).
\item The waypoints are connected
by a G2 clothoid spline (corresponding to the solid blue line of
Figure~\ref{fig:pathFitting}), using the algorithms and techniques
discussed in ~\cite{bertolazzi2018g2, bertolazzi2018interpolating},
and for which an efficient C++ implementation is
available~\cite{bertolazzi2018clothoids}.
\end{enumerate}
The choice of the clothoid
comes from the observation that humans tend to follow a unicycle-like
dynamics~\cite{farina2017walking} given in~\eqref{eq:Ekf2}, which
naturally generates clothoid curves. What is more, clothoids have been
proved to be effective to mimic a human path by a robotic
agent~\cite{BevilacquaFFP18ral}, for the continuity of the curves and
of their curvature.

\subsection{Robot control}
When a path is reconstructed, following the steps described above, the
controller module takes the responsibility to execute a safe
navigation of the robot following as closely as possible the
prescribed path.  For this work, we employed the path following
algorithm described in~\cite{andreetto2017harnessing}, which is
velocity-independent and avoids the singularities presented by other
common algorithms when the vehicle has to stop and the velocity is set
to zero.  The velocity of the robot is chosen by our controller based
both on the distance from the end of the path (corresponding to the
leader position), with the aim of following the leader at a constant
(curvilinear) distance, on the current path curvature (the vehicle is
slowed down when traveling a sharp curve), and on the past robot
velocities (to limit the maximum allowed accelerations).

In addition to following the path, the control module implements a
safety policy whereby when an obstacle is encountered along the path
the robot first slows down, and then stops if the occlusion does not
pass away.  During this phase, if the distance from the leader becomes
too large, the robot emits a sound signal to notify the exception and
to attract the leader's attention.

%% file: experiments.tex
\section{Experimental results}
\label{sec:experiments}

A first set of experiments were to decide the most effective
combination of solutions for vision-based detection and
recognition. As regards detection, our evaluation lead to the adoption
of SSD, since it implements the CNN with a relatively small number of
parameters. The result is a low computation time, which comes at the
price of a slight detection inaccuracy, which is however compensated
by the fusion module. The recognition module based on KNN
has been tested on the Market1501 dataset~\cite{zheng2015scalable}: by
comparing the results obtained by the KNN in combination with ResNet50
(which produced approximately $50\%$ of correct responses) and
GoogLeNet (more than $90\%$ of correct responses), the latter emerged
as an obvious choice. Finally, for the image tracker, our aim is to
process long real-time sequence with occasional total occlusions and
changes of shape.  We evaluated the computation performance of each of
the methods presented in Section~\ref{sec:related}. The results, which
are reported in Table~\ref{table:tracking-comparison}
frame-per-second, suggested us the adoption of KCF, CSRT and MOSSE.
\begin{table}[t]
\centering
\caption{Overview of the FPS rate of the image tracking algorithms. The
  performances were measured on an Intel Core i5 CPU and on an Nvidia
  Jetson TX2 GPU.}
\label{table:tracking-comparison}
	\begin{tabularx}{\columnwidth}{Xccccccc}
	\toprule
	& MIL & KCF & MedFlow & CSRT & MOSSE & GOTURN & TLD \\
	\midrule
	FPS & 9 & 38 & 40 & 15 & 56 & 20 & 10 \\
	\bottomrule
	\end{tabularx}
\end{table}

\subsection{Performance evaluation of the system as a whole}
\label{subsec:follower}

\begin{figure}[t]
  \centering
  \begin{tabular}{cc}
    \includegraphics[width=0.4\columnwidth]{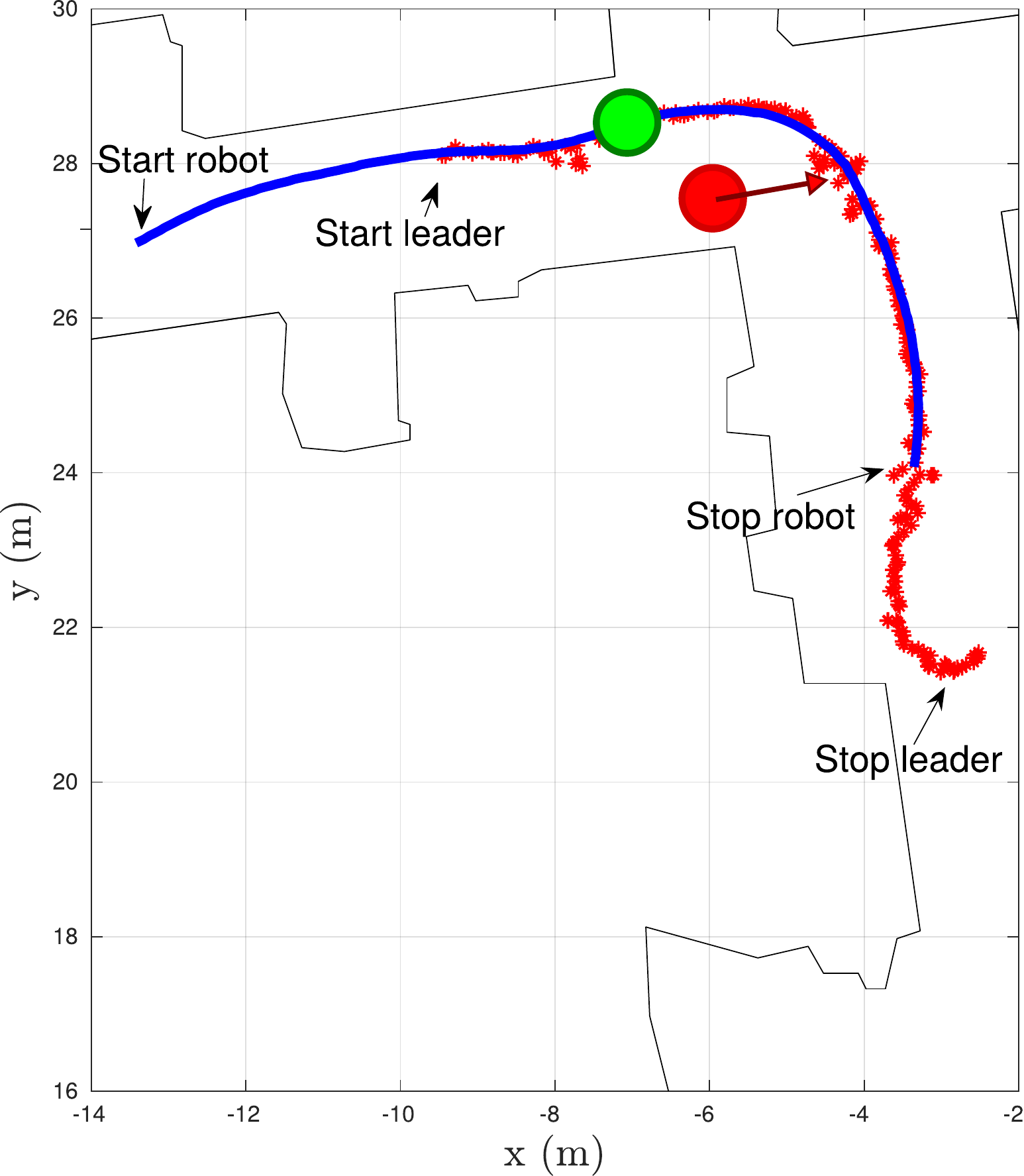} &
                                                        \includegraphics[width=0.45\columnwidth]{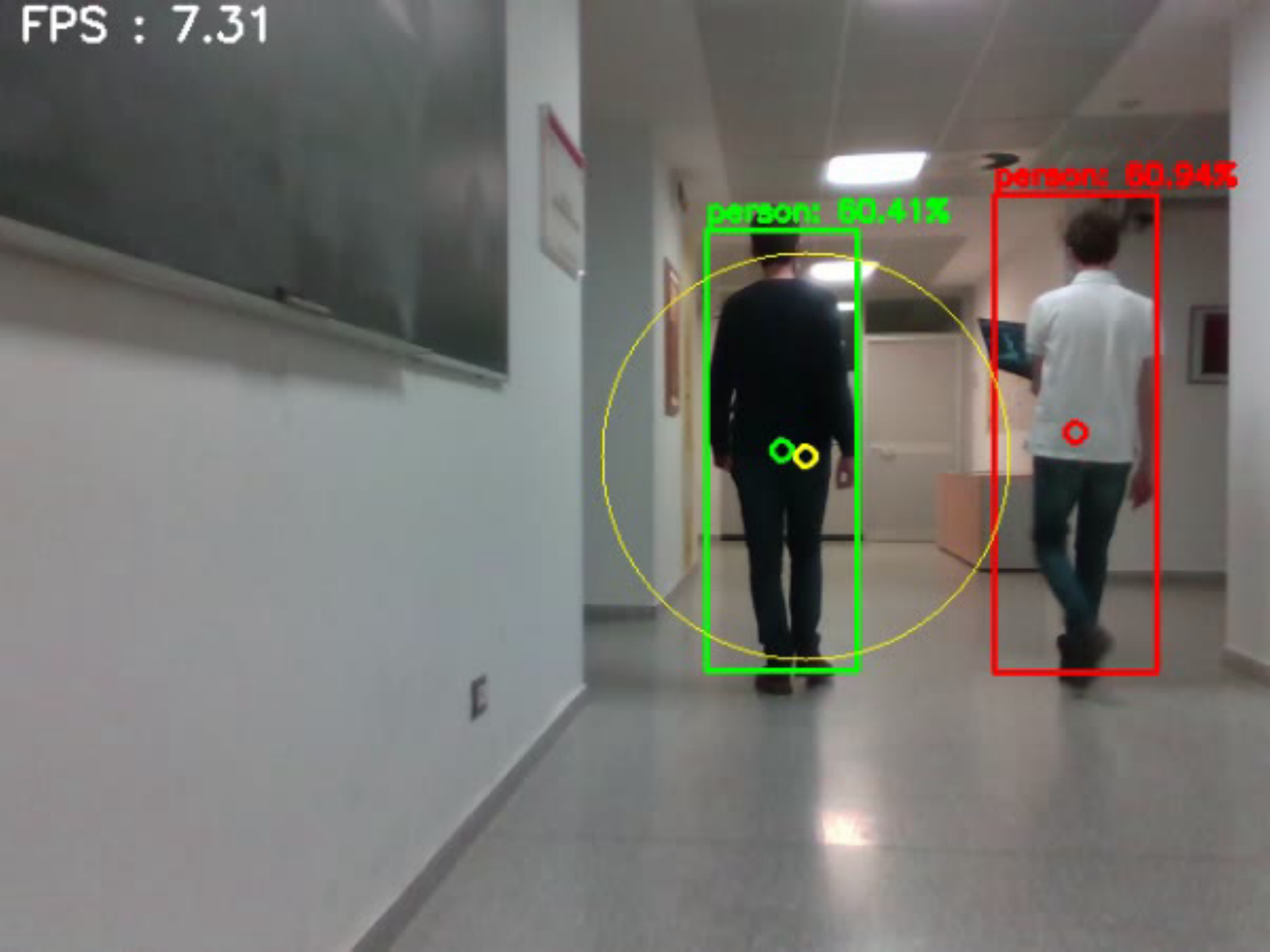} \\
    (a) & (b) \\
    \includegraphics[width=0.4\columnwidth]{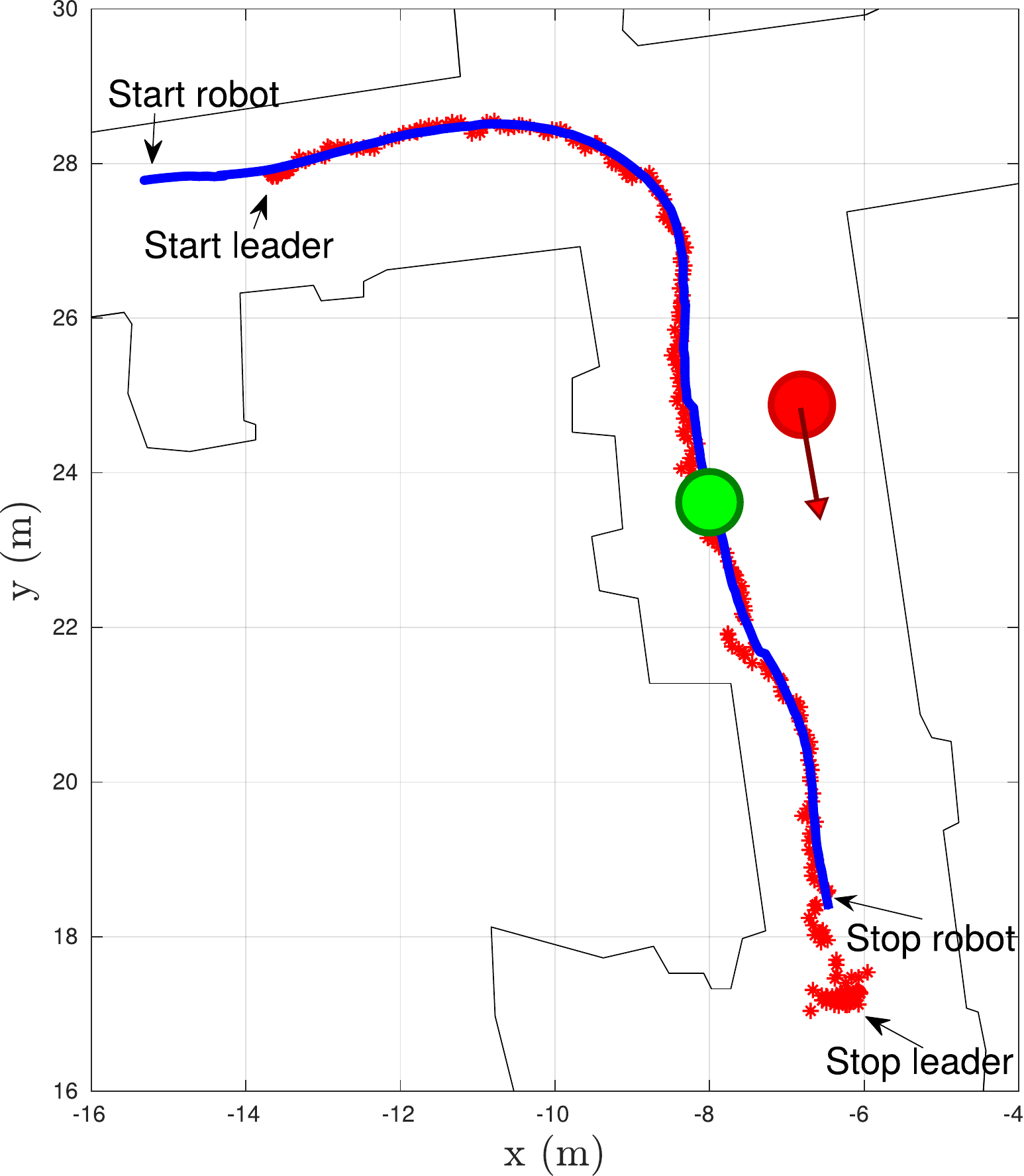} &
                                                        \includegraphics[width=0.45\columnwidth]{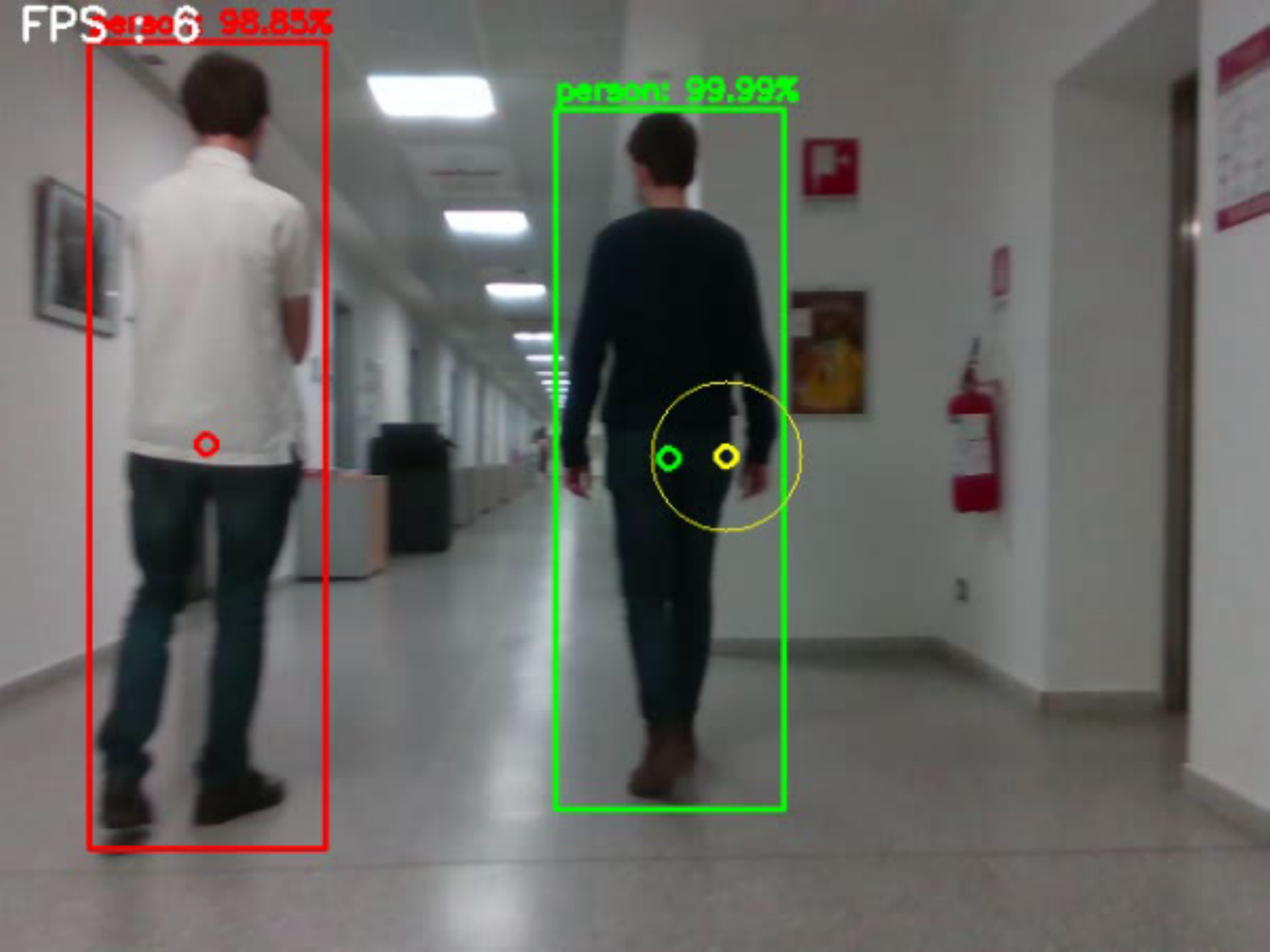} \\
    (c) & (d) \\
  \end{tabular}
  \caption{Experimental trajectories in a hallway. The leader and a
    pedestrian walk in the same corridor with (a,b) partially
    occluding trajectories and (c,d) missing and recovering of the
    leader with the camera tracking. (a,c) depict the trajectory
    followed by the robot (solid blue line) and the measured leader
    positions (red stars), while the green and red circles correspond
    to the positions of the leader and the other pedestrian,
    respectively, when the camera snapshots shown in (b,d) are
    grabbed.}
  \label{fig:experiment-elevators}
\end{figure}

The algorithms presented in Section~\ref{sec:architecture} and
Section~\ref{sec:navigation} were executed on a Jetson
TX2\footnote{https://www.nvidia.com/en-us/autonomous-machines/embedded-systems/jetson-tx2/}
for the acquisition of the RGB-D data and the classificator, while the
LIDAR scans, the sensing data fusion, and the navigation control were
executed on a NUC, both on board of the wheeled robot entirely assembled
at the University of Trento. Furthers experiments were carried out in
our department at the University of Trento. First, we present the
performance and robustness of the leader tracking algorithm.  To this
end, we record the data in two different portions of an hallway of our
department with multiple exits and in different circumstances. In
Figure~\ref{fig:experiment-elevators}-b, the robot follows the leader
while another pedestrian is walking nearby, however the tracking is
correctly maintained on the leader (see
Figure~\ref{fig:experiment-elevators}-a for the trajectories).
Similar results are obtained for crossing trajectories or when the
leader exits from the camera field of view for the right turn but the
other pedestrian is not wrongly classified as the leader, which is
tracked back after the turn
(Figure~\ref{fig:experiment-elevators}-c,d).

\begin{figure}[t]
  \centering
  \includegraphics[width=0.9\columnwidth]{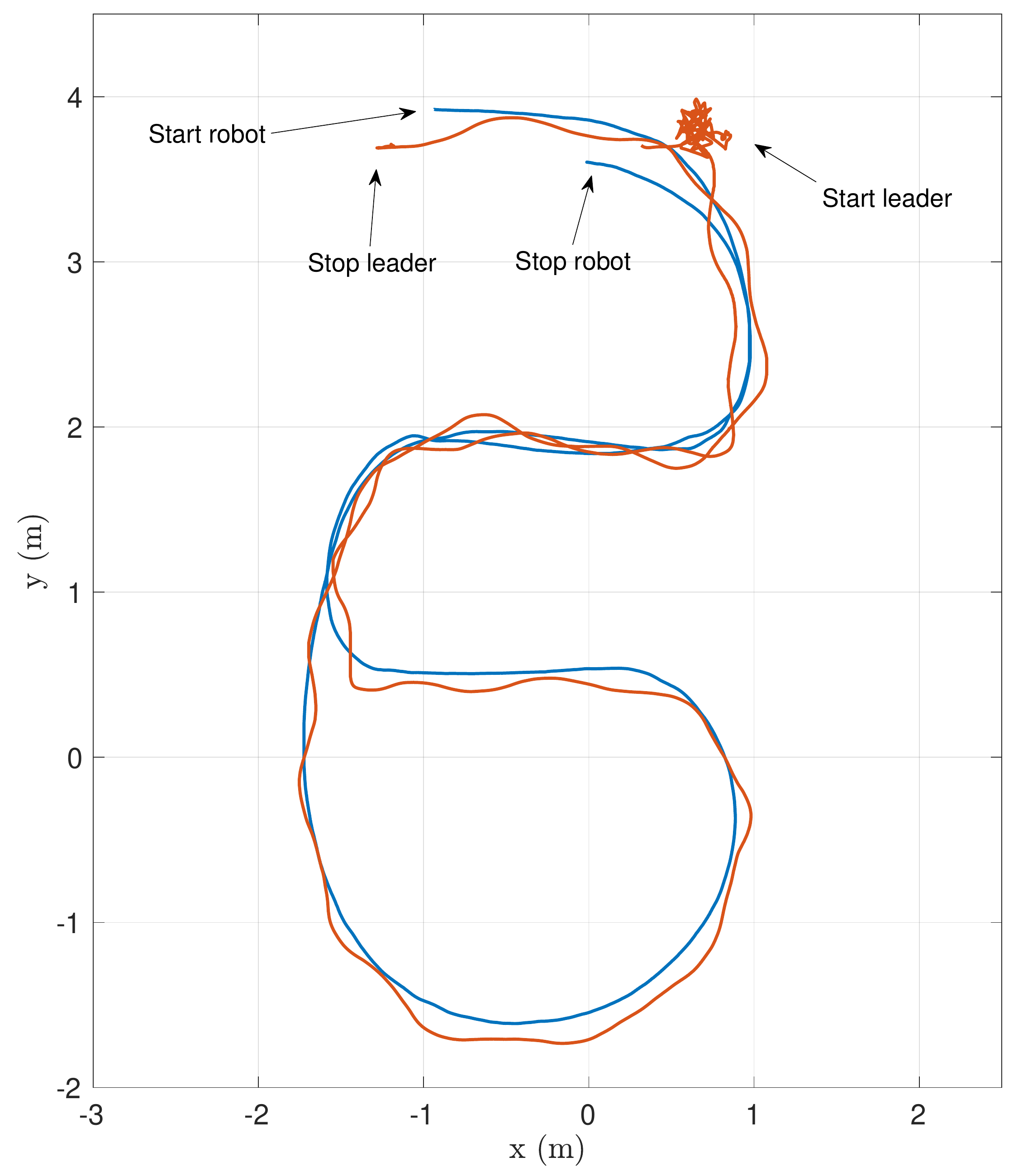}
  \caption{An example of the actual leader path (red line) and robot
    trajectory (blue line).}
  \label{fig:experiment-optitrack}
\end{figure}

For a qualitative analysis of the tracking and navigation, we present
in Figure~\ref{fig:experiment-optitrack} an example of the comparison
of the robot trajectory (blue line) with the actual position of the
leader (red line), both captured with a network of eight OptiTrack
cameras for ground truth reference.
Notice the leader starting position standing in front of the robot
during the bootstrap phase. The swinging leader trajectory is dictated
by the OptiTrack tracked markers placed on the head of the human to
avoid occlusions, hence oscillating with the footsteps. From this
picture it is evident that, in sharp turns, the robot looses the image
tracking of the leader, but it is nonetheless able to exactly
follow his path by means of the fusion with the LIDAR
data. Finally, we would like to point out that the error in the
trajectory followed by the robot with respect to the human footsteps
is in the range of $\pm 25$~cm, i.e. the typical encumbrance of the
human body.

Due to stringent space limitations, further experimental evidences of
the effectiveness of the approach can be found in the video
accompanying this paper.

%% file: conclusions.tex
\section{Conclusions and future work}
\label{sec:conclusion}

In this paper, we have presented an approach for guiding a robot
across a difficult environment. A human operator takes the role of a
path-finder and the robot follows, moving in a close neighbourhood of
the path physically marked by the human with her/his footsteps. This
application required a combination of state-of-the-art techniques for
robust perception and path reconstruction.  The experimental results
show the high level of reliability and robustness reached by the
proposed solution.

Different points remain open and are reserved for future work. A first
work direction is releasing the assumption that the robot is in
possession of a map and can rely on a global localisation system. We
envisage a scenario in which the map is constructed while the robot
follows its leader and it only relies on a relative localisation with
respect to reference points identified in the environment and to its
leader. Another important direction is a theoretical study of how the
interaction between model based approaches and neural networks can
produce results with a guaranteed accuracy.  Finally, a possibility we
are considering is the use of wearable haptic bracelets and the
implementation of a protocol that the robot can use to notify to its
leader the occurrence of exceptional conditions (e.g., when the path is
too close to an obstacle and the robot cannot follow it within
appropriate safety margins).